\documentclass[11pt]{article}
\usepackage[T1]{fontenc}
\usepackage{lmodern}
\usepackage[margin=1in]{geometry}
\usepackage{amsmath,amssymb,graphicx,booktabs,array,url,xcolor,float,microtype,cite}
\usepackage[hidelinks]{hyperref}
\hypersetup{
  pdftitle={HyperDAM: Hyperspectral Distractor-Aware Memory with Amodal Expansion for SAM 3 Tracking},
  pdfauthor={Ryoga Yuzawa and Tasuku Takagi}
}

\newcommand{\method}{HyperDAM}
\newcommand{\dataset}{HOTC2026-Modal}
\newcommand{\extendedcite}[2]{\cite{#1,#2}}
\newcommand{\extendedtext}[1]{#1}
\newcommand{\versiontext}[2]{#2}
\newcommand{\fitwide}[1]{\begingroup\hfuzz=100pt\resizebox{\textwidth}{!}{#1}\endgroup}
\newcommand{\figureonewidth}{0.6\textwidth}
\newcommand{\tabletwosize}{\footnotesize}
\newenvironment{paperfigure}{\begin{figure}[H]}{\end{figure}}
\newenvironment{paperfigurewide}{\begin{figure}[H]}{\end{figure}}
\newenvironment{papertable}{\begin{table}[H]}{\end{table}}
\newenvironment{papertablewide}{\begin{table}[H]}{\end{table}}
\newcommand{\appendixheadingbreak}{\texorpdfstring{\\}{ }}

\title{HyperDAM: Hyperspectral Distractor-Aware Memory with Amodal Expansion
for SAM 3 Tracking}

\author{Ryoga Yuzawa \qquad Tasuku Takagi\thanks{E-mail:
  \texttt{ryoga.yuzawa at gmail.com, tasuku.takagi at gmail.com}}}
\date{}

\begin{document}
\maketitle
\pagestyle{plain}

\begin{abstract}
Hyperspectral video provides material cues that can disambiguate targets with
similar false-color appearance, yet foundation-model trackers update memory
primarily from spatial and appearance evidence. We present \method{}, a
DAM4SAM3-based hyperspectral tracker with three principal contributions.
First, \dataset{} adds human-verified frame-wise modal masks and mask-tight boxes
to all 481 organizer-provided HOTC 2026 videos. Second, a frame-zero-calibrated
HSI gate rejects spectrally inconsistent updates to the distractor-resolving
memory (DRM) without altering the current prediction. Third, a causal spatiotemporal expander adds
outward-only amodal corrections from frozen SAM features. Static-scene recovery
and empty-mask RTS smoothing address target switches and full occlusion. Model
selection prioritizes cross-domain robustness over leaderboard-specific optimization.
The final system ranked second in HOTC 2026, achieving 68.0093\% AUC and
87.7703\% DP@20 in the organizer's private evaluation.

\end{abstract}

\begin{center}
distractor-aware memory; amodal tracking; modal annotations; spectral identity
\end{center}

\section{Introduction}
\label{sec:introduction}

Single-object tracking estimates a target trajectory from an initial bounding
box. Hyperspectral video augments conventional appearance with densely sampled
spectral responses, providing evidence about material identity when color or
texture alone is ambiguous. In camouflaged scenes, a target and its background
may have nearly identical visual appearances while retaining distinct spectral
signatures, making spectral evidence essential when appearance is unreliable
\extendedcite{wang2025bihot}{nguyen2010reflectance,xiong2020material,chen2024sense}.
Existing hyperspectral benchmarks and trackers demonstrate
the value of spectral--spatial modeling
\extendedcite{xiong2020material,liu2022h3,chen2024phtrack,he2025sphst}{uzkent2018deepkcf,li2020baenet,liu2021hanet,li2021sstat,wang2022ssatfn,wang2023hsptrack,liu2024tsssiamese,wang2024ssfnet}, while limited
training data and the spectral-band gap continue to favor adaptation of strong
RGB trackers
\extendedcite{wang2025hotsurvey}{muszynski2023helios,xie2023vphot,wan2024spatracker,mohamed2024spectraltransformer}.

Promptable segmentation models provide a complementary path. \versiontext{The
Segment Anything family supports transferable segmentation from visual prompts
\cite{ravi2024sam2,carion2026sam3}, and}{SAM introduced promptable image
segmentation from visual prompts \cite{kirillov2023sam}. SAM~2 extended this
paradigm to video using streaming memory \cite{ravi2024sam2}, while SAM~3
unified detection, segmentation, and tracking from concept and visual prompts
\cite{carion2026sam3}.} DAM4SAM improves
tracking robustness by managing distractors in a dedicated memory
\cite{videnovic2025dam4sam}. Two gaps remain in the HOTC setting. First, an
incorrect high-confidence mask can still enter memory and contaminate future
predictions. Second, a segmentation mask describes only the visible part of an
occluded target, whereas the required tracking box may extend beyond the
visible support.

Figure~\ref{fig:amodal-rts} previews how the proposed output trajectory handles
the latter gap. The learned expander restores latent extent while the target is
partially visible, and RTS supplies a box only when the SAM mask is absent.

\begin{paperfigure}
  \centering
  \includegraphics[width=\figureonewidth]{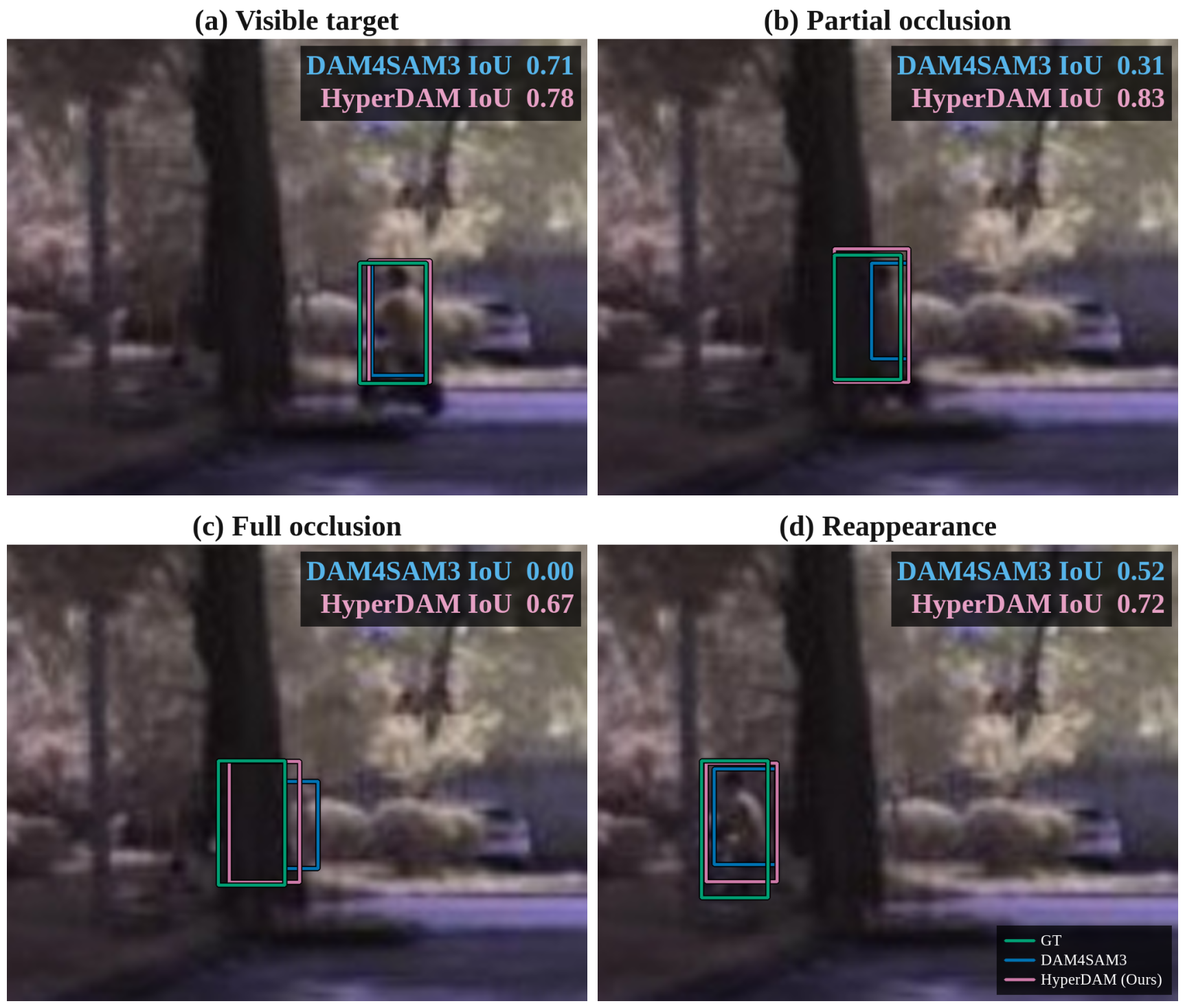}
  \caption{HyperDAM through an occlusion on \texttt{nir-rider6}. From left to
  right and top to bottom, the target is visible, partially occluded, fully
  occluded, and visible again. The amodal expander corrects partial visibility,
  while empty-mask RTS bridges full occlusion. Green, blue, and pink denote
  ground truth, DAM4SAM3, and HyperDAM, respectively.}
  \label{fig:amodal-rts}
\end{paperfigure}

We address these gaps with \method{}, a system that separates three responsibilities:
spectral evidence protects target identity, a learned head estimates latent
object extent, and the underlying frozen tracker owns the base trajectory. Our
contributions are:

\begin{itemize}
  \item \textbf{A reviewed modal-annotation extension.} We augment the
  organizer-provided HOTC 2026 competition dataset with binary visible/modal
  masks and mask-tight boxes for the 406 training/update videos and the
  organizer's 75-video public-leaderboard set (denoted Public-LB75).

  \item \textbf{A hyperspectral memory gate.} A sequence-local model calibrated
  from the frame-zero target evaluates material consistency at DAM4SAM's native
  distractor-resolving memory (DRM) update boundary. It can reject a requested update but cannot request one
  or alter the current-frame output.

  \item \textbf{An amodal expander for DAM4SAM3 boxes.} A causal six-frame
  3-D head trained from paired modal masks and challenge boxes predicts
  nonnegative per-side expansion. The expansion is transferred as a residual
  onto the HSI-controlled trajectory and is never fed back into SAM memory.
\end{itemize}

Static-scene recovery and empty-mask Rauch--Tung--Striebel (RTS) smoothing
\cite{rauch1965maximum} are supporting components rather than
primary contributions.
They are included to specify the submitted system completely.
The implementation and reproducibility artifacts are publicly available at
\url{https://github.com/RyogaYuzawa/hotc2026-hyperdam}.

\begin{paperfigurewide}
  \centering
  \includegraphics[width=\textwidth]{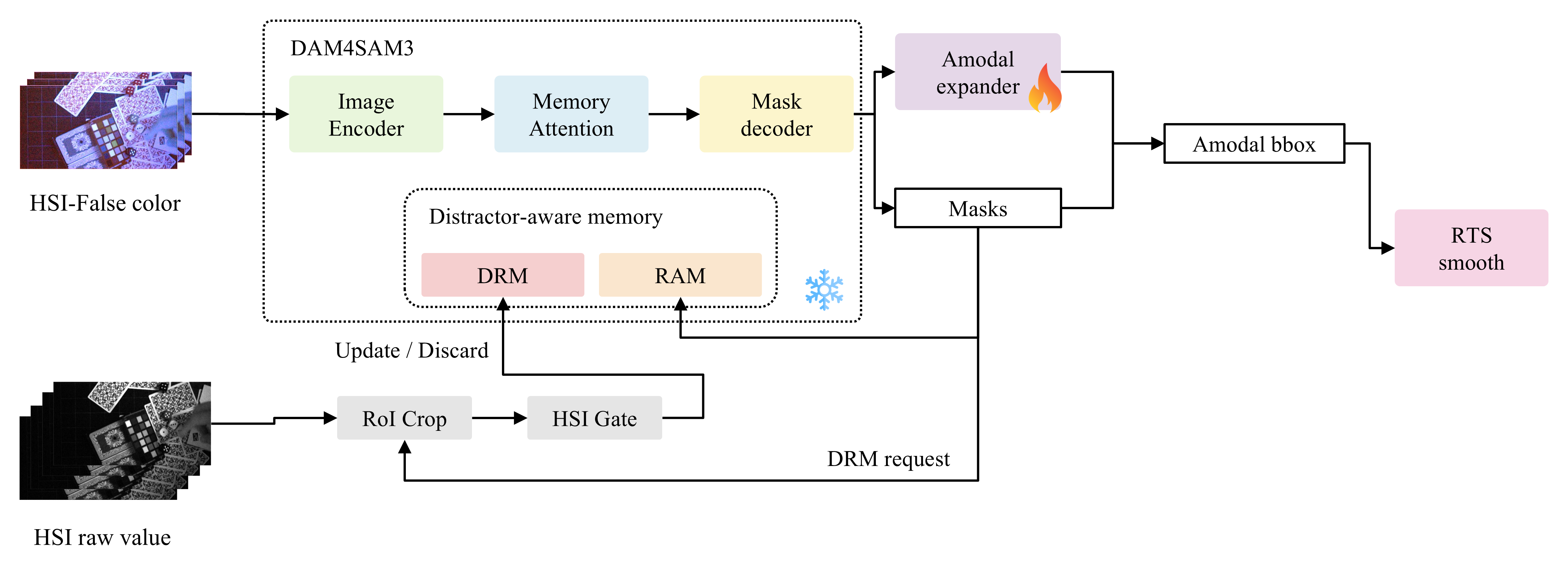}
  \caption{Overview of the proposed HyperDAM architecture. The HSI gate
  filters requested DRM updates using raw hyperspectral observations;
  DAM4SAM3 supplies modal masks and features for amodal expansion, followed by
  offline RTS smoothing.}
  \label{fig:architecture}
\end{paperfigurewide}

\section{Related Work}
\label{sec:related}

\subsection{Hyperspectral object tracking}

Hyperspectral tracking has progressed from material descriptors and
correlation filters to learned spectral--spatial representations
\extendedcite{xiong2020material,liu2022h3,wang2025bihot}{nguyen2010reflectance,uzkent2018deepkcf,li2020baenet,liu2021hanet,li2021sstat,wang2022bssiamrpn,su2022threebranch,wang2022ssatfn,wang2023hsptrack,liu2024tsssiamese}.
Recent methods transfer RGB
trackers or pretrained representations through material-aware fusion and
spectral prompts
\extendedcite{chen2024phtrack,he2025sphst}{liu2023multiband,muszynski2023helios,xie2023vphot,chen2024sense,wang2024ssfnet,wan2024spatracker,chen2024hysstp,wang2024promptmamba,mohamed2024spectraltransformer}.
WHISPERS entries have
explored detection-driven SAM2, low-rank adaptation, SAM2-based HOT, and
motion-aware memory
\cite{he2024desam2,chen2024lahst,qian2024hypersam,chen2025harmony}.
\extendedtext{Other systems combine SAM~2 with depth and Kalman-based
reinitialization for amodal tracking \cite{medellin2024amodal}, adapt frozen RGB
trackers through prompting and adaptive token dropping \cite{zhou2024adtrack},
or use detector-driven self-prompting and zero-shot SAM~2 tracking
\cite{yang2025detection,zhang2025sam2hot}. PGSR-Track introduces position
guidance with bidirectional spatial--spectral attention \cite{li2025pgsr}, while
SAM2Local fuses SAM~2 with a local tracker and check-and-reprompt logic
\cite{lai2025sam2local}. }
\versiontext{Our
spectral component instead acts as a sequence-local, label-free decision layer:
it does not retrain the tracker and intervenes only when the native tracker
requests a DRM update.}{Our spectral component targets a specific decision in
memory management: when DAM4SAM3 requests a DRM update, a sequence-local,
label-free spectral test can reject that write. The test leaves the native
update schedule, the current output, and the tracker weights unchanged;
it neither injects learned feature prompts nor replaces the tracker's
feature extractor.}

\subsection{Foundation models for segmentation and tracking}

SAM~2 introduced promptable video segmentation with streaming memory
\cite{ravi2024sam2}, while SAM~3 unified detection, segmentation, and tracking
from concept and visual prompts \cite{carion2026sam3}. DAM4SAM augments the
video tracker with a distractor-aware memory and an introspection-based update
policy \cite{videnovic2025dam4sam}. Our HSI gate tests material identity at
DAM4SAM's final DRM update boundary.

\subsection{Modal and amodal supervision}

Modal annotations describe visible pixels, while amodal annotations represent
the complete object extent behind occluders. TAO-Amodal shows that heavy
occlusion remains challenging for modern trackers
\cite{hsieh2024taoamodal}.
\extendedtext{Kar et al. infer amodal bounding boxes from visible regions in
single images \cite{kar2015amodal}. MOVi-MC-AC provides synthetic multi-camera
amodal-completion data \cite{moore2025xray}, while Amodal SAM extends promptable
segmentation to recover hidden object extent \cite{zhang2026amodalsam}.
Medellin et al. combine SAM~2, depth, and Kalman-based reinitialization for
amodal hyperspectral tracking \cite{medellin2024amodal}. }
Our extension adds dense modal masks and mask-tight
boxes to the organizer-provided HOTC 2026 competition videos. The expander
learns the difference between this visible extent and the challenge-provided
target box, while retaining a strict outward-only correction contract.

\section{HOTC 2026 Dataset Extension}
\label{sec:dataset}

HOTC 2026 provides 406 training/update videos and the 75-video Public-LB75
split with paired hyperspectral observations. The former have frame-wise amodal target
boxes, whereas leaderboard ground truth is withheld. We extend all 481 videos
with human-verified visible/modal segmentation masks and their mask-tight
boxes, and release these annotations as \dataset{}. This annotation layer
supplements rather than replaces the organizer-provided data. Paired with the
official hyperspectral and false-color frames, the masks and boxes can supervise
spectral adaptation of segmentation models, domain adaptation to false-color
imagery, and losses that distinguish modal observations from amodal target
extent. The \dataset{} annotation release is available at
\url{https://huggingface.co/datasets/ryo818/HOTC2026-Modal}.
\versiontext{Existing HOT
benchmarks are predominantly box-annotated and remain modest in scale
\extendedcite{wang2025bihot}{liu2022h3,xiong2020material,wang2025hotsurvey}; our extension targets the
missing dense modal supervision rather than introducing a new tracking split.}{Existing HOT benchmarks support box-based tracking evaluation
\cite{wang2025bihot,xiong2020material,wang2025hotsurvey}.
Our extension adds dense modal supervision to the organizer-provided videos
rather than introducing a new tracking split.}

DAM4SAM3 predicts a visible mask and hence a modal box, while HOTC supervision
specifies the object's amodal extent. Directly using the latter as a mask or
visible-box target conflates observed pixels with hidden extent. Our paired
modal masks and boxes separate these quantities, enabling the amodal expander
to learn outward residuals from modal to amodal boxes while SAM3 and the base
tracker remain frozen.

SAM3 first infers and propagates target masks. Annotators visually inspect its
outputs and manually correct erroneous frames or regions; only reviewed masks
enter the release. Each modal box is then computed as the tight axis-aligned
box around visible foreground pixels. An empty mask denotes full occlusion.
The release also records official frame identifiers, available challenge
boxes, provenance, and checksums.

The release contains a 406-video primary split and a separate 75-video
Public-LB75 annotation-only split. Because official leaderboard ground truth is not
released, the latter annotations are not presented as challenge ground truth
and are excluded from fitting, model selection, and component evaluation.

\section{Method}
\label{sec:method}

The proposed system uses frozen DAM4SAM3 as its base tracker and augments it
with three principal components. First, an HSI-gated memory mechanism controls
whether a requested DRM update is committed,
preventing spectrally inconsistent objects from contaminating the tracker
state. Second, an amodal expander enlarges the native modal box toward the
estimated full extent of a partially occluded object. Third, an offline
Rauch--Tung--Striebel (RTS) smoother fills fully occluded frames for which the
tracker produces no valid mask. We additionally exploit the prevalence of
fixed-view cameras in the dataset by classifying each scene as static or
dynamic. In static scenes, abrupt and spatially asymmetric box growth is
treated as a likely tracking error and triggers static-scene recovery.
The HSI gate is the only component that modifies the DRM update policy.
Static-scene recovery may reinitialize the base tracker, whereas amodal
expansion and RTS smoothing operate only on the output trajectory and never
feed back into DAM4SAM3 memory. Figure~\ref{fig:architecture} summarizes the
complete inference pipeline.

\subsection{DAM4SAM3 base tracker}

Our base tracker adapts the distractor-aware DAM4SAM memory design
\extendedcite{videnovic2025dam4sam}{ravi2024sam2,carion2026sam3} to frozen SAM3.\extendedtext{ We use the
DAM4SAM3 implementation provided by SAM3-TrackBench
\cite{alansari2025memory}, pinned to revision
\texttt{d37e4a975e48}. The bundled runtime and its checksum are recorded in
the released artifact manifest.} It runs in PVS-only mode with
detector inference disabled and is initialized by the official frame-zero box.
SAM3 propagates the target mask and exposes alternative mask hypotheses;
DAM4SAM3 uses these outputs to manage recent appearance memory and
DRM. The tight box of the selected mask is the
native modal trajectory. We use the entire base tracker in frozen form, with
all SAM3 parameters unchanged. We initially explored \versiontext{LoRA adaptation using the
HOTC training data \cite{hu2022lora}}{LoRA adaptation~\cite{hu2022lora}
using the HOTC training data}, but the limited size and diversity of
the dataset did not yield the expected generalization to held-out sequences.
We therefore retain the frozen model and instead introduce the HSI-based
memory gate described below.

\subsection{HSI gate}
\label{sec:hsi-gate}

DAM4SAM3 first decides when to request a DRM update. For each request, the HSI
gate evaluates the proposed update mask
$M_t$ together with an expanded surrounding region
$E_t=\operatorname{Dilate}(M_t)\setminus M_t$. First, the raw sensor mosaic
$I_t^{\mathrm{raw}}$ is decoded into an aligned HSI cube
$X_t\in\mathbb{R}^{H\times W\times B}$:
\begin{equation}
 X_t=\mathcal{D}(I_t^{\mathrm{raw}}).
 \label{eq:hsi-cube}
\end{equation}
Here $B$ is the number of spectral bands: $B=16$ for VIS, $B=25$ for
NIR, and $B=15$ for RedNIR.
Let $p_0=\operatorname*{median}_{x\in M_0}X_0(x)$ be the reference spectrum
from the frame-zero target mask. For each requested DRM update, the component-wise
median spectra of the candidate mask and expanded region are
\begin{equation}
\begin{aligned}
 p_t^M&=\operatorname*{median}_{x\in M_t}X_t(x),\qquad
 p_t^E=\operatorname*{median}_{x\in E_t}X_t(x),\\
 s_t^M&=\cos\!\left(\widehat{p_t^M},\widehat{p_0}\right),\qquad
 s_t^E=\cos\!\left(\widehat{p_t^E},\widehat{p_0}\right).
\end{aligned}
 \label{eq:spectral-similarity}
\end{equation}
Here $\widehat{\cdot}$ denotes mean centering across spectral bands followed
by $\ell_2$ normalization. The DRM update is accepted only when the candidate
is similar to the initial target and its similarity exceeds that of the
expanded region:
\begin{equation}
\begin{aligned}
 A_t^{\mathrm{HSI}}&=\mathbb{1}\!\left[
 s_t^M\geq\tau_{\mathrm{id}}\ \land\
 s_t^M-s_t^E\geq\tau_{\mathrm{exp}}\right],\\
 W_t&=U_t\land A_t^{\mathrm{HSI}},
\end{aligned}
  \label{eq:write-gate}
\end{equation}
Here $U_t$ is DAM4SAM3's native DRM update request. The thresholds are
calibrated from the frame-zero target and its expanded region. The gate may
reject the requested DRM update but cannot request an update itself; rejection
leaves both the current output and DRM unchanged.

No cross-sequence training or ground-truth annotation is used by the HSI gate.
An unreliable frame-zero reference abstains rather than rejecting.

Figure~\ref{fig:hsi-gate} shows an example. At $t=124$, DAM4SAM3 requests and
performs a DRM update, whereas the HSI gate rejects it because the candidate
mask does not separate sufficiently from its surroundings. The current boxes
remain unchanged, but the different memory states produce the later divergence
at $t=236$--$237$.

\begin{paperfigurewide}
  \centering
  \includegraphics[width=\textwidth]{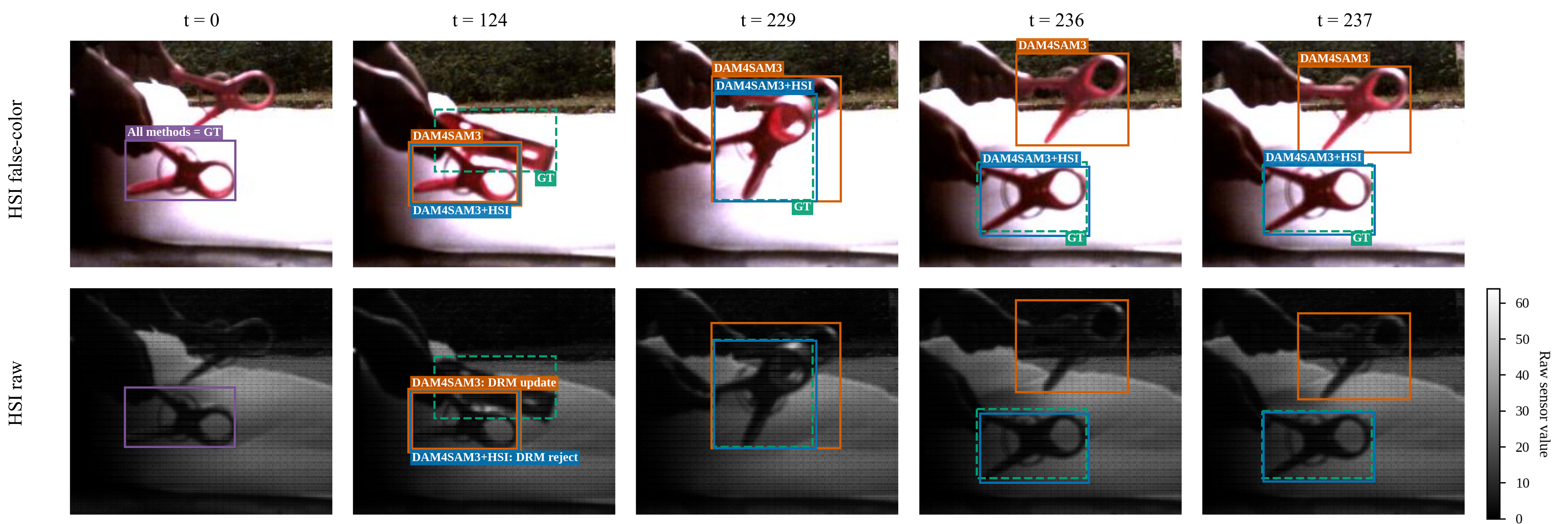}
  \caption{Runtime trace of the HSI gate on \texttt{vis-clamp2} from the
  HOTC 2026 dataset. The rows show
  aligned false-color frames and raw 16-band mosaics. At $t=124$, DAM4SAM3
  accepts the DRM update, whereas DAM4SAM3+HSI rejects it without changing
  the current output. By $t=236$--$237$, the native tracker follows the
  visually similar distractor while the gated tracker remains on the target.
  Orange, blue, and green dashed boxes denote DAM4SAM3, DAM4SAM3+HSI, and
  ground truth, respectively.}
  \label{fig:hsi-gate}
\end{paperfigurewide}

\subsection{Amodal expander for SAM}
\label{sec:amodal}

The SAM family predicts the visible segmentation mask of a target, and its
bounding box is therefore the tight envelope of that modal mask. HOTC instead
evaluates the amodal extent, including portions hidden by occlusion, so a
SAM-based tracker requires an explicit modal-to-amodal adaptation. Motivated
by the plug-in expander in TAO-Amodal
\cite{hsieh2024taoamodal}, we preserve all
pretrained SAM3 weights and attach an external amodal expander head to the
detached outputs of the DAM4SAM3 mask decoder. Only this head is trained to
infer outward box corrections for the hidden extent.

For a SAM mask box $B_t=(x_t,y_t,w_t,h_t)$, the expander predicts nonnegative
outward residuals $\Delta_t=(\delta_{\mathrm L},\delta_{\mathrm T},
\delta_{\mathrm R},\delta_{\mathrm B})$ in pixel
units. The amodal box is obtained by directly adding these side residuals:
\begin{equation}
\begin{aligned}
 \operatorname{Amodal}(B_t;\Delta_t)=\big(&
 x_t-\delta_{\mathrm L},\ y_t-\delta_{\mathrm T},\\
 &w_t+\delta_{\mathrm L}+\delta_{\mathrm R},\
 h_t+\delta_{\mathrm T}+\delta_{\mathrm B}\big).
\end{aligned}
 \label{eq:amodal-output}
\end{equation}
Training targets are the outward differences between the frozen SAM box and
the official amodal box. A side is supervised as positive only when the
reviewed modal box indicates genuine occlusion, preventing expansion when SAM
already reaches or overshoots the official boundary.

The expander is causal and temporal. From the final 32-channel feature map of
the SAM3 mask decoder, it extracts a box-aligned $32\times24\times24$ crop at
each frame and stacks the latest six crops. Aligned mask logits, the
256-dimensional object query, box geometry, and an immutable frame-zero anchor
are supplied alongside this feature history. Six residual 3-D convolutional
blocks aggregate the spatiotemporal information, after which four side heads
predict the left, top, right, and bottom residuals in Eq.~\eqref{eq:amodal-output}.
Confidence gates set unsupported side corrections to zero.

The external head has 9.4M parameters and is pretrained on synthetic
MOVi-MC-AC amodal data~\cite{moore2025xray}. All SAM3 parameters remain frozen.
Frames with full occlusion are excluded from expander training, which therefore
targets partial occlusion where a visible SAM box can be expanded toward the
amodal box.

\subsection{Full-occlusion handling}

The amodal expander moves partially occluded boxes toward the amodal
ground-truth extent, but learning reliable boxes under full occlusion is
difficult because no visible mask remains. We therefore handle full
occlusion after the full video has been processed. A frame is marked as a
missing observation when the independent frozen DAM4SAM3 pass emits an empty
mask; the released pipeline does not apply an additional confidence threshold.
Rauch--Tung--Striebel (RTS) smoothing \cite{rauch1965maximum} then reconstructs
only these missing boxes from valid observations before and after the gap,
leaving all observed boxes unchanged.

Figure~\ref{fig:amodal-rts} illustrates the complementary roles of the two
components. The amodal expander recovers the hidden extent while part of the
target remains visible, whereas RTS smoothing supplies the box when the target
is fully occluded and the SAM mask is empty.

\subsection{Static/dynamic scene classification}

Our analysis found that many HOTC sequences use a fixed camera and that target
switches in such scenes are often accompanied by an abrupt, spatially
asymmetric change in box size. We therefore estimate background motion from
\versiontext{sparse trackable points over the initial ten frames
\extendedcite{shi1994goodfeatures}{lucas1981iterative}, then classify each sequence as static or dynamic.}{Shi--Tomasi corners~\cite{shi1994goodfeatures} tracked with pyramidal
Lucas--Kanade optical flow~\cite{lucas1981iterative} over the initial ten
frames, then classify each sequence as static or dynamic.}

\section{Experiments}
\label{sec:experiments}

\subsection{Contribution of each component}

Table~\ref{tab:ablation} reports a cumulative ablation on the 75-video
Public-LB75 split (26,860 frames) using the same
frozen DAM4SAM3 checkpoint and inference revision. Every ablation row was
submitted to the organizer's public Kaggle evaluation. The final
private-leaderboard result is included in the same table for reference.

\begin{papertable}
  \centering
  \caption{Cumulative ablation on Public-LB75.}
  \label{tab:ablation}
  \small
  \begin{tabular*}{\columnwidth}{@{\extracolsep{\fill}}clc@{}}
    \toprule
    ID & Configuration & Public score (AUC) \\
    \midrule
    B0 & DAM4SAM3 native & 0.69039 \\
    B1 & B0 + HSI gate & 0.69214 \\
    B2 & B1 + static-scene recovery & 0.69513 \\
    B3 & B2 + amodal expander & 0.69732 \\
    B4 & B3 + RTS smoothing & \textbf{0.69768} \\
    \midrule
    \multicolumn{2}{@{}l}{Organizer's private LB} &
      \shortstack{AUC: 68.0093\% \\ DP@20: 87.7703\%} \\
    \bottomrule
  \end{tabular*}
\end{papertable}

\begin{papertablewide}
  \centering
  \caption{Unified benchmark on Public-LB75 (75 videos; 26,860 frames).
  Public AUC is evaluated on the Kaggle public leaderboard. Available
  peak-VRAM measurements use the same NVIDIA L4 GPU. The best AUC value is
  shown in bold.}
  \label{tab:main-benchmark}
  \tabletwosize
  \fitwide{\begin{tabular*}{\textwidth}{@{\extracolsep{\fill}}lcrrcp{0.13\textwidth}@{}}
    \toprule
    Method & Public-LB75 AUC $\uparrow$
      & Peak VRAM (MiB) $\downarrow$
      & Params (M) $\downarrow$
      & Precision
      & Comment \\
    \midrule
    OSTrack \cite{ye2022ostrack} & 0.49512 & 443.24 & 92.83 & FP32 & -- \\
    SeqTrack \cite{chen2023seqtrack} & 0.54142 & 1601.95 & 308.98 & FP32 & -- \\
    HIPTrack \cite{cai2024hiptrack} & 0.57374 & 557.88 & 120.41 & FP32 & -- \\
    ODTrack \cite{zheng2024odtrack} & 0.54442 & 599.99 & 92.83 & FP32 & -- \\
    SUTrack \cite{chen2025sutrack} & 0.57169 & 2205.59 & 746.61 & FP32 & -- \\
    \addlinespace[1pt]
    SAM2 \cite{ravi2024sam2}
      & 0.65041 & 1634.95 & 224.45 & FP32 & -- \\
    DAM4SAM \cite{videnovic2025dam4sam}
      & 0.65464 & 2177.24 & 224.45 & FP32 & SAM2 base\newline algorithm \\
    SAM3 \cite{carion2026sam3}
      & 0.67393 & 4530.28 & 827.31 & FP32 & PVS \\
    DAM4SAM3 \cite{videnovic2025dam4sam,carion2026sam3}
      & 0.69039 & 4492.51 & 827.31 & FP32 & PVS \\
    \midrule
    \method{} (Ours)
      & \textbf{0.69768} & 4685.12 & 836.73 & FP32 & -- \\
    \bottomrule
  \end{tabular*}}
\end{papertablewide}

\subsection{Benchmark}

Table~\ref{tab:main-benchmark} reports all verified results on the same
Public-LB75 protocol. Each method is initialized with the provided first-frame
box and produces one prediction per frame. All models are evaluated in FP32 on
a single NVIDIA L4 GPU. Although \method{} was selected with cross-domain
generalization rather than leaderboard optimization as the primary objective,
it achieves the highest AUC among the compared methods and improves upon its
DAM4SAM3 base. This result indicates that the proposed components improve the
base tracker without sacrificing performance on the public evaluation domain.

Public-LB75 ground truth is withheld after the initial box and Kaggle returns
only AUC. We therefore do not import published values obtained with different
sequence sets or metrics into this benchmark.
Table~\ref{tab:main-benchmark} also gives peak allocated VRAM and parameter
count where available.

\subsection{Discussion}

The gains in Table~\ref{tab:ablation} reflect complementary roles: the HSI gate
prevents visually similar distractors from contaminating memory
(Figure~\ref{fig:hsi-gate}), static-scene recovery handles target switches, the
amodal expander handles partial occlusion, and RTS fills full-occlusion gaps.
Their individually modest gains therefore accumulate across distinct failure
modes.

\subsection{Limitations}

The HSI gate yields only a modest gain because it only rejects requested DRM
updates. Future work could initialize an object detector and associate multiple
candidate spectral prototypes with the target by HSI correlation, moving
material reasoning from memory validation to candidate selection.

\section{Conclusion}
\label{sec:conclusion}

We presented \method{}, extending frozen DAM4SAM3 with a frame-zero HSI update
gate, outward-only amodal expansion, static-scene recovery, and empty-mask RTS,
together with \dataset{} annotations for all 481 HOTC 2026 videos. The system
achieved the best Public-LB75 AUC among the evaluated methods while
cross-domain selection limited leaderboard overfitting. Its organizer-private
scores were 68.0093\% AUC and 87.7703\% DP@20, ranking second in HOTC 2026.
Future work will explore detector-assisted spectral association and online
trajectory completion.

\bibliographystyle{IEEEbib}
\bibliography{refs-arxiv}

\clearpage
\appendix
\section{Dataset Composition Across HOTC406 and the\appendixheadingbreak
Public-Leaderboard Set}
\label{app:dataset-composition}

We audit the dataset composition using only sequence titles and frame counts,
without inspecting withheld leaderboard ground truth. A title family is formed by
lowercasing a sequence title and removing only its terminal numeric suffix;
thus, for example, \texttt{car3} and \texttt{car11} belong to \texttt{car},
whereas context-bearing names such as \texttt{high\_car} remain distinct.
The resulting families are then assigned to the broad semantic groups in
Table~\ref{tab:title-family-map}. This title-derived taxonomy is intended to
describe dataset composition rather than provide ground-truth object labels.

Table~\ref{tab:title-family-map} summarizes all 406 organizer-provided
training/update videos (167,724 frames) and the separate 75-video public-
leaderboard set (Public-LB75; 26,860 frames). HOTC406 is dominated by vehicles,
people, and everyday
objects, whereas Public-LB75 places substantially more weight on vegetation,
food, and animals. The two splits contain 110 and 31 normalized title families,
respectively; only six names occur in both: \texttt{car}, \texttt{dashcam},
\texttt{dronecam}, \texttt{pills}, \texttt{pingpong}, and \texttt{truck}.

\begin{table}[H]
  \centering
  \caption{Complete assignment of normalized title families to broad
  categories. Numeric title variants within a listed family are grouped
  together. V and F are the total numbers of videos and frames, respectively,
  in each category and split.}
  \label{tab:title-family-map}
  \scriptsize
  \setlength{\tabcolsep}{2.5pt}
  \renewcommand{\arraystretch}{1.05}
  \begin{tabular}{@{}p{0.14\textwidth}p{0.31\textwidth}rrp{0.23\textwidth}rr@{}}
    \toprule
    & \multicolumn{3}{c}{HOTC406} & \multicolumn{3}{c}{Public-LB75} \\
    \cmidrule(lr){2-4}\cmidrule(l){5-7}
    Category & Title families & V & F & Title families & V & F \\
    \midrule
    Vehicles and traffic
      & automobile, bus, car, dashcam, drive, excavator, high\_car,
        high\_truck, l\_car, rainystreet, surround\_car, taxi, toy\_car, truck
      & 137 & 33,829
      & bicycle, car, dashcam, motorcycle, rccar, truck
      & 21 & 6,234 \\
    People and human activity
      & face, foot, hand, high\_person, high\_rider, l\_basketball\_person,
        l\_person, l\_runner, pedestrain, pedestrian, people, player, rider,
        s\_jump, s\_person, s\_runner, s\_walker, s\_warmup, student,
        surround\_person, trucker, worker
      & 65 & 26,593
      & athlete, head, walker
      & 4 & 1,266 \\
    Sports and balls
      & badminton, ball, ball\_holder, basketball, football, high\_playground,
        l\_basketball, pingpong, playground, pool, s\_soccer, s\_volleyball,
        snow\_table\_tennis, snow\_tennis
      & 43 & 23,120
      & pingpong
      & 4 & 1,114 \\
    Animals
      & ant, duck, kangaroo, turkey
      & 10 & 4,508
      & bee, cat, dog
      & 7 & 3,373 \\
    Vegetation and food
      & apple, coke, cranberries, fake\_orange, forest, fruit, grove, leaf,
        leaves, oranges, real\_pear
      & 30 & 17,712
      & dryleaf, flower, folium, herbs, moss, setaria, vegetation
      & 15 & 6,433 \\
    Toys and tabletop games
      & ball\&mirror, card, cards, dice, mirror\_egg, rubber\_duck, rubik,
        snow\_card, toy, yo\_yo
      & 39 & 20,760
      & doll, domino, marble, marblerun, poker
      & 12 & 3,872 \\
    Everyday objects
      & backpack, balloon, board, book, bracelet, clamp, cloth, coin, cup,
        glass, glass\_cup, hat, keyboard, officechair, officefan, paper,
        paper\_crane, paper\_frog, pen, pills, plastic\_cups, receipts, redbag,
        whitecup
      & 55 & 30,771
      & clip, earphone\_case, headset, pills
      & 8 & 3,188 \\
    Medical scenes
      & esophagectomy, heartsurgery
      & 3 & 1,191
      & ---
      & 0 & 0 \\
    Aerial and illuminated scenes
      & drone, dronecam, droneshow, partylights
      & 18 & 6,147
      & dronecam
      & 1 & 200 \\
    Other scenes and objects
      & bytheriver, campus, park, shadow, stone
      & 6 & 3,093
      & jelly
      & 3 & 1,180 \\
    \midrule
    Total & 110 families & 406 & 167,724 & 31 families & 75 & 26,860 \\
    \bottomrule
  \end{tabular}
\end{table}

\end{document}